# Real-Time Apple Detection System Using Embedded Systems With Hardware Accelerators: An Edge AI Application

VITTORIO MAZZIA[1,2,3], (Student Member, IEEE),
ALEEM KHALIQ[1,2], (Student Member, IEEE),
FRANCESCO SALVETTI[1,2,3], (Student Member, IEEE),
AND MARCELLO CHIABERGE[1,2], (Member, IEEE)
[1]Department of Electronics and Telecommunications, Politecnico di Torino, 10129 Turin, Italy
[2]Politecnico di Torino Interdepartmental Centre for Service Robotics (PIC4SeR), 10129 Turin, Italy
[3]SmartData@PoliTo—Big Data and Data Science Laboratory, 10129 Turin, Italy

Corresponding author: Vittorio Mazzia (vittorio.mazzia@polito.it)

This work was supported by the PoliTO Interdepartmental Centre for Service Robotics (PIC4SeR) and SmartData@Polito at the Politecnico di Torino, Italy.

**ABSTRACT** Real-time apple detection in orchards is one of the most effective ways of estimating apple yields, which helps in managing apple supplies more effectively. Traditional detection methods used highly computational machine learning algorithms with intensive hardware set up, which are not suitable for infield real-time apple detection due to their weight and power constraints. In this study, a real-time embedded solution inspired from "Edge AI" is proposed for apple detection with the implementation of YOLOv3-tiny algorithm on various embedded platforms such as Raspberry Pi 3 B+ in combination with Intel Movidius Neural Computing Stick (NCS), Nvidia's Jetson Nano and Jetson AGX Xavier. Data set for training were compiled using acquired images during field survey of apple orchard situated in the north region of Italy, and images used for testing were taken from widely used google data set by filtering out the images containing apples in different scenes to ensure the robustness of the algorithm. The proposed study adapts YOLOv3-tiny architecture to detect small objects. It shows the feasibility of deployment of the customized model on cheap and power-efficient embedded hardware without compromising mean average detection accuracy (83.64%) and achieved frame rate up to 30 fps even for the difficult scenarios such as overlapping apples, complex background, less exposure of apple due to leaves and branches. Furthermore, the proposed embedded solution can be deployed on the unmanned ground vehicles to detect, count, and measure the size of the apples in real-time to help the farmers and agronomists in their decision making and management skills.

## I. INTRODUCTION

Monitoring agricultural farms and orchards mainly rely on skilled farmers and workers who are responsible for assessing several growth stages before perform-farming related actions in order to maximize the quality and yield. Manual work of these farmers consumes time and increases production costs, and workers with less knowledge and experience make unnecessary mistakes. With the advancements in precision agriculture and information technology, crop imaging has become an important source of information that can be used to assess vegetation status of the crops, fruits growth, yield, and quality.

Two important features that enable the farmers to estimate crop-load and yield mapping in tree fruit crops are fruit counting and size estimation. Several studies have proposed fruit detection in orchards using machine vision systems for automatic growth assessment, robotic harvesting, and yield estimation [1], [2]. Apple crop-load management has gained much importance due to its impact on yield production. It has been the primary problem to develop algorithms that enable the apple harvesting robot to directly, quickly, and accurately recognize fruits in real-time [3]. In the natural environment, for the visual systems, apple fruit detection is typically more difficult because of the influence of lights and shadows, branches, and leaf coverings. Apple's visual appearances in the natural environment may be categorized as non-occluded fruits and occluded fruits.

Occlusion of fruits due to leaves, branches, and other fruits and variable lighting conditions are some of the main reasons that make it more challenging to achieve good accuracy and robustness in fruit detection [1]. Experiments in few studies have been performed in nighttime environments with the formation of tunnel structures around tree canopies to deal with variable lighting conditions. Images from both front and back ends are taken of tree canopies using multiple sensors to avoid fruit occlusion, [4], [5], [7], which leads to having high fruit detection accuracy. Nevertheless, fruit size estimation is needed for automatic robotic harvesting of mature and good-sized fruits, due to the difficulty level of the real-time robotic harvesting, few studies [7], [8], have exploited fruit size estimation using a machine vision system. Wang *et al.* [7] performed experiments using an RGB-D camera a thin lens theory to estimate the size of mango fruits in trees. Ultrasonic sensors have also been tested with the color images for size estimation of citrus fruit along with the range information Regunathan *et al.* [9].

With the upsurge of machine learning, deep learning algorithms have been extensively used in agriculture-related applications [10]. Deep learning can be used for crop mapping [11], [12], crop image segmentation [13], crop target detection [14], [15]. Convolutional Neural Networks (CNNs) are used in [16] to extract target regions in the image, object segmentation, and counting number of fruits on a tree using a successive CNN counting algorithm. Dias *et al.* [13] used CNN in combination with support vector machine (SVM) to extract the features of apple blossoms automatically way to counter complex background, which leads to achieving comparatively accurate apple blossom area segmentation results than the previous studies. Faster R-CNN [17] was employed with the region proposal network (RPN) method to detect the region of interest (ROI) in the image with a complex background scene followed by a classifier, which classifies bounding boxes. Faster R-CNN with VGG16 net [18] is the state-of-art method in fruit detection [10]. However, Faster R-CNN consists of region proposal networks (RPN) and classification networks that produced excellent results in terms of accuracy, while the detection speed is slow, which can not achieve good results in real-time with high image resolution. The You Only Look Once (YOLO) method [19], [20] deals with the classification and the localization as a regression problem. A YOLO network directly performs regression to detect targets in the image without RPN, hence it is fast and can be implemented in real-time applications. The state-of-art version (YOLOv3) [20] not only has high detection accuracy and speed but also performs well with detecting small targets. However, the YOLOv3 model is not suitable for real-time applications such as in harvesting robots due to its complex architecture that requires more processing power. Optimization of the parameters of the model reduces the computational complexities and thus is needed to deploy on edge devices such as Jetson, and Raspberry Pi.

Large data sets training and validation require high-performance computing machines such as clusters or servers, which are widely being used in deployment of power extensive deep learning algorithms [21], [22], however, in the low power end devices, researchers have raised their concern about efficiency of CNNs, in real-time embedded platforms [23], [24]. Network optimization (i.e., network pruning or quantization) is a technique to reduce the model size by compressing the dense model into sparse or low-bit architecture with minimal or even no accuracy drops.

### A. AI ON THE EDGE AND RELATED WORK

Real time smart solutions inspired from deep learning, must possess the following key capabilities such as energy efficient, affordable and small form factor with the fine balance between accuracy and power consumption. Indeed, deep learning based architecture are conventionally deployed with in the centralized cloud computing environment. However, there are constraints such as considerable latency of the network, energy and financial overheads that effects the overall performance of the system. To deal with these limitations, edge computing often called ''edge AI'' has been introduced where computations are performed locally on the data acquired from various devices or sensors.

The challenge in meeting the implementation requirements for edge AI is to ensure high output accuracy of algorithms while consuming low power. Nevertheless, the innovation in hardware options, involving central processing units (CPUs), graphics processing units (GPUs), application-specific integrated circuits (ASICs), and system-on-a-chip (SoC) accelerators, has made edge AI possible. NVIDIA, Intel, and Qualcomm are the leading market brands which are contributing enormously to the development of AI at the edge. Among these, Intel's Movidius Neural Computing Stick (NCS) is the cheapest device to implement computationally extensive algorithms with multiple layers of CNN. In [25], CNN model was deployed in NCS to perform classification of 3D voxel based point clouds.

NVIDIA's Jetson is another uprising embedded hardware and broadly used accelerators for machine learning algorithms. Promising feature of Jetson is the CPU-GPU heterogeneous architecture [26], [27], where CPU boots up the firmware and the CUDA-capable GPU come with the potential to accelerate complex machine-learning tasks. Key features includes form factor, light weight and low power consumption. However, to gain full potential of Jetson and attaining real-time performance involves optimization phase to both Jetson hardware and NN algorithms. Jetson variants termed as TK1, TX1, and TX2 are widely used in past few years. For example, in [28], low cost TK1 was used in drowsiness detection using model compression of deep neural networks. Nvidia TX1 was used in tennis ball collection robot based on deep learning [29]. In [30], qualitative comparison was made among various hardware platforms, TX2 ranked the highest in terms of throughput. They used Tiny-YOLO for object detection and claimed better product of accuracy and frame rate than YOLO and SSD. [31] deployed Tiny-YOLO on TX2 to perform detection and localization of the robot using Kinect-V2 visual sensor. Casecaded CNN model was deployed in TX2 for semantic weed classification using multi spectral images for smart farming [32].

In our work, we deployed a modified version of the YOLOv3-tiny algorithm on embedded platforms such as Raspberry Pi 3 B+ in combination with Intel Movidius Neural Computing Stick (NCS), Nvidia's Jetson Nano and Jetson AGX Xavier for real-time apple detection.

The rest of the paper is organized as follows. Section 2 will cover the data set description and hardware details. In section 3, the architecture framework is described with further explanation of the performed customization for small object detection. Section 4 will describe the experimental results and discussion followed by the conclusion.

## II. MATERIALS AND DATA

An orchard has been considered in order to acquire a custom data set for the training process. Subsequently, a technique dubbed transfer learning [47] has been applied, re-training, and fine-tuning a custom version of the network YOlOv3-tiny specifically optimized for accurate detection of small objects on embedded devices. After training, the resulting network has been benchmarked on all images of OIDv4 [46] data set (training, validation, and testing) with the Apple class, producing a reproducible metric that can be easily compared with future works. Finally, the trained model has been tested on several edge AI devices assessing their performance in terms of speed and power consumption.

### A. DATA SET DESCRIPTION

Two popular types of apple (Braeburn and Fuji) were considered in this study, which are the most common types found in the north part of Italy. Image acquisition campaign was conducted in randomly selected healthy apple trees in orchards using a reflex digital camera with 18 megapixels during different times and days of September. Image acquisition was performed for separate/non-overlapped fruits, overlapping fruits/occluded fruits under variable lighting conditions such as fully exposed to sun from front, full sun influencing from the back of the fruits, and fruits covered by the shades of leaves/branches or other apples.

### B. HARDWARE DESCRIPTION

The concept of Edge AI consists of performing computations locally on an embedded system in real-time. Since the training process requires a lot more computational power as compared to the inference process, it is not performed on the embedded system, but a dedicated workstation. Then, the model with the obtained weights is deployed on the target hardware in order to be executed.

The workstation used for training was equipped with an NVIDIA RTX 2080Ti GPU with CUDA 10 and 64GB of DDR4 SDRAM. This GPU model features 544 Tensor Cores, an NVIDIA technology specifically designed to boost matrix multiplication performance, thus able to speed up the training process of deep learning models. The computational power of this GPU allows reaching peak performances of about 26.9 TFLOPs (FP16) [33].

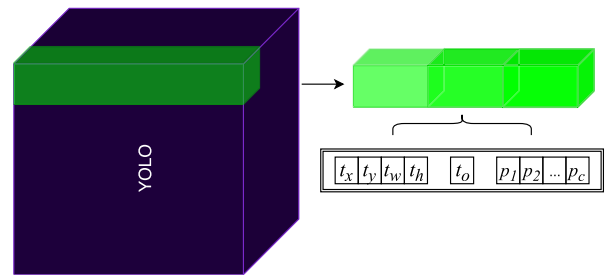

**FIGURE 1.** The 1x1 convolution predicts, for each location of the first two dimensions of the input tensor, an array with [3 ∗ (5 + C)] where C is the number of classes. So, with COCO data set the output tensor is encoded with a dimension 255.

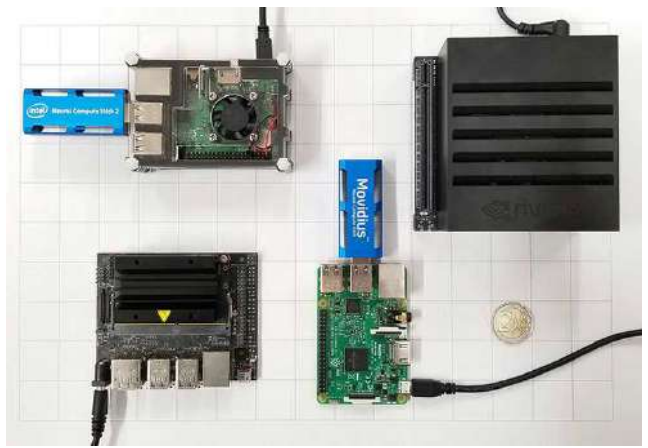

**FIGURE 2.** The analyzed embedded platforms. Top left: Raspberry Pi 3 B+ with Intel NCS2. Top right: NVIDIA Jetson AGX Xavier. Bottom left: NVIDIA Jetson Nano. Bottom right: Raspberry Pi 3 B+ with Movidius NCS.

For the embedded implementation of the model, different hardware platforms have been considered, as shown in Fig. 2.: a Raspberry Pi 3 B+ with both generation of Intel Movidius Neural Compute Stick accelerators, an NVIDIA Jetson Nano, and an NVIDIA Jetson AGX Xavier. Table. 1. shows a comparison between the main specifications of the selected embedded hardware.

Neural Compute Sticks (NCS) are USB hardware accelerators specifically designed to perform AI computations. Two generations of NCS have been tested: the first is powered by a Myriad 2 VPU (Visual Processing Unit) processor [34], while the second features a Myriad X VPU [35]. These two chips are designed by Intel Movidius to accelerate deep neural network inferences. Since the Neural Sticks provide a USB 3.0 interface, they are suitable to be used with embedded, lightweight, and cheap computers such as a Raspberry.

NVIDIA, on the other hand, provides a family of boards that feature an embedded computer with a dedicated GPU

TABLE 1. Main specifications of the platforms investigated in this study. Each device is reported with its related commercial price at the time of publication. It is important to point out that the two versions of Intel Movidius Neural Compute Stick (NCS) necessitate an external embedded system. For our research, we exploited a Raspberry Pi 3 B+.

|  | Intel Movidius NCS | Intel NCS 2 | Jetson Nano | Jetson AGX Xavier |
| --- | --- | --- | --- | --- |
| Features size | 73 x 26 mm | 73 x 26 mm | 70 x 45 mm | 100 x 87 mm |
| HW Accelerator | Myriad 2 VPU | Myriad X VPU | 128-core NVIDIA Maxwell GPU | 512-core NVIDIA Volta GPU with 64 Tensor Cores and 2x NVDLA Engines |
| CPU | N.A. | N.A. | Quad-core Arm A57 @ 1.43 GHz | Octa-core NVIDIA Carmel Arm |
| Memory | 4GB LPDDR3 | 4GB LPDDR3 | 4GB LPDDR4 | 16GB LPDDR4x |
| Storage | N.A. | N.A. | Micro SD card slot or 16 GB eMMC flash | 32 GB eMMC 5.1 |
| Peak performance | 100 GFLOPs | 150 GFLOPs | 472 GFLOPs | 16 TFLOPs |
| Native precision support | FP16 | FP16 | FP16 / FP32 | FP16 / FP32 |
| Nominal power | 1 W | 1.5 W | 5 / 10 W | 10 / 15 / 30 W |
| Weight | 18 g | 19 g | 140 g | 280 g |
| Price | $ 70 | $ 74 | $ 99 | $ 800 |

for hardware acceleration. The boards examined in this work, the AGX Xavier and the Nano, are the last two Jetson platforms presented by NVIDIA.

The AGX Xavier has been released in Autumn 2018 and currently is the most powerful Jetson board available. It features a Volta GPU micro-architecture with 64 Tensor Cores, able to reach up to 11 TFLOPs (FP16), and two NVDLA (NVIDIA Deep Learning Accelerator) engines. These chips are specifically designed to perform neural network standard operations such as convolutions efficiently. A single NVDLA is able to compute up to 2.5 TFLOPS. Thus, the overall peak performance of the AGX Xavier is about 16 TFLOPS. The board can work in different power modes, and it gives the user the possibility to select the number of working CPU cores. The available power modes are 10W (2 cores), 15W (4 cores), 30W (2, 4, 6 or 8 cores) [36].

The Jetson Nano has been presented in June 2019 especially for target applications where reducing the board size, power consumption, and price is important. For the hardware acceleration, it features an NVIDIA Maxwell GPU with a peak performance of 472 GFLOPs. The Nano board does not include any deep learning specific accelerator and can work in two power modes at 5W or 10W [36]. So, it does not take advantage of Tensor cores and NVDLA engines for inference acceleration.

## III. METHODOLOGY AND ARCHITECTURE FRAMEWORK

YOLO is a network specifically designed for fast and accurate real-time object detection. It has comparable performance in terms of accuracy with other popular object detection algorithms like RetinaNet [37], Faster-RCNN [38], but it is much faster and compact that makes it an optimal choice for real-time embedded applications. It is a single fully convolutional neural (FCN) network that takes as input a raw image and gives as output bounding boxes and related classes of recognized objects inside the presented scene.

Since 2016 different versions have been released [19], [20], [39] that gradually have increased the accuracy of the general framework without giving away too much of its inference speed. At the same time, all different versions have been released with a lighter counterpart dubbed ''tiny'' that has a simplified and optimized structure without loss of too much accuracy. The intrinsic characteristics of the ''tiny'' version make it suitable for AI applications at the edge with the use of embedded systems, enhanced with hardware accelerators. For this reason, this research has taken the last available ''tiny'' version of YOLO, YOLOv3-tiny, as a starting point for the realization of an embedded apple detector system.

In the rest of this section, fundamental working principles of the network and the modifications applied to the original ''tiny'' architecture are presented in order to make it suitable for the detection of smaller objects in the scene like an apple.

### A. ARCHITETURE OF THE ORIGINAL FRAMEWORK

YOLOv3-tiny, as already introduced, makes use of only convolutional layers, making it a fully convolutional network that can accept inputs of different sizes during and after training. It can be divided into two main blocks: the first one is the feature extractor or backbone dubbed darknet-19. Its principal and the fundamental role are to extract features in a hierarchical fashion a starting from raw pixels coming from the input layer. Indeed, the extracted representations are later used as starting point by the other modules of the network. Darknet-19 is a light and efficient feature extractor, but can be easily swapped with any other backbone like ResNet [40], DenseNet [41], etc. It features a standard architecture greatly inspired by VGGNet [18], making use of only 3x3 filters throughout the entire structure, max-pooling layers in order to reduce the dimensionality of the input volume and obtain local invariance. Finally, darknet-19 exploits Batch Normalization layers [43] to accelerate the network training, reducing the internal covariance shift. All backbone blocks use

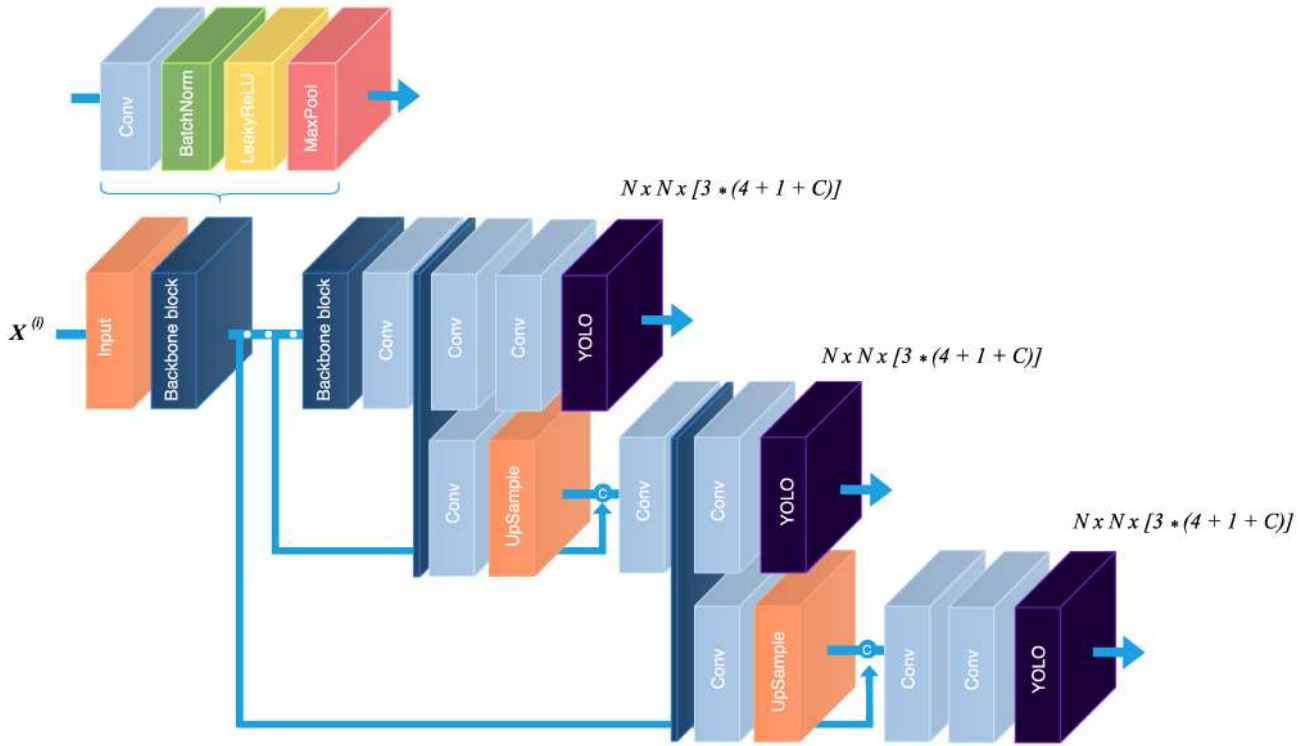

**FIGURE 3.** Overview of the modified version of the model YOLOv3-tiny. The original architecture features only the two first branches because it makes detection at two different scales instead of three. The detection layers exploits feature maps of three different sizes, having strides 32, 16, 8. Each YOLO layer predicts for each cell three bounding boxes using three different anchors.

LeackyReLU [42] as activation function. On the other hand, the second block of the YOLOv3-tiny architecture analyzes the produced backbone representations, and it predicts position and class of belonging of the different objects present in the input raw image. That is achieved with 1x1 convolutions arranged in a Pyramid Network [44] structure. The structure of the original architecture is unrolled in Table. 2. where each layer is presented with its output tensor dimension.

### B. INTERPRETING THE PREDICTIONS

Given an input image, the final output prediction of the network is a list of bounding boxes along with the recognized classes. First of all, the backbone reduces dimension of input images by a factor called the stride of the network. Then, the features extracted by the cascade of convolutional layers feed a classifier/regressor, which makes the detection predictions. That is performed by a 1x1 convolutional layer always placed before the "YOLO" layer. So, the output prediction of the network is a feature map tensor that has the same first two dimensions of the previous layer.

For example, settings of the network shown in Table. 2., after layer 14, the following 1x1 convolution processes the input tensor of dimension 13x13x512 and produces for each 13x13 location a vector with 255 elements. In Fig. 1. is depicted how these arrays are internally arranged. Each of them can be divided into $B$ sections, where each section is responsible for predicting a specific bounding box.

**TABLE 2.** YOLOv3-tiny original architecture with input images of dimension 416×416. In this example, the two 1×1 convolutional layers 15 and 22 that make the final predictions are set to work with the COCO [45] data set (80 different classes).

| Layer | Type | Size/Stride | Filters | Output |
|---|---|---|---|---|
| 0 | Conv | 3 x 3/1 | 16 | 416 x 416 x 16 |
| 1 | MaxPool | 2 x 2/2 | | 208 x 208 x 16 |
| 2 | Conv | 3 x 3/1 | 32 | 208 x 208 x 32 |
| 3 | MaxPool | 2 x 2/2 | | 104 x 104 x 32 |
| 4 | Conv | 3 x 3/1 | 64 | 104 x 104 x 64 |
| 5 | MaxPool | 2 x 2/2 | | 52 x 52 x 64 |
| 6 | Conv | 3 x 3/1 | 128 | 52 x 52 x 128 |
| 7 | MaxPool | 2 x 2/2 | | 26 x 26 x 128 |
| 8 | Conv | 3 x 3/1 | 256 | 26 x 26 x 256 |
| 9 | MaxPool | 2 x 2/2 | | 13 x 13 x 256 |
| 10 | Conv | 3 x 3/1 | 512 | 13 x 13 x 512 |
| 11 | MaxPool | 2 x 2/1 | | 13 x 13 x 512 |
| 12 | Conv | 3 x 3/1 | 1024 | 13 x 13 x 1024 |
| 13 | Conv | 1 x 1/1 | 256 | 13 x 13 x 256 |
| 14 | Conv | 3 x 3/1 | 512 | 13 x 13 x 512 |
| 15 | Conv | 1 x 1/1 | 255 | 13 x 13 x 255 |
| 16 | YOLO | | | |
| 17 | Route 13 | | | |
| 18 | Conv | 1 x 1/1 | 128 | 13 x 13 x 128 |
| 19 | Up-samp | 2 x 2/1 | | 26 x 26 x 128 |
| 20 | Route 19 8 | | | |
| 21 | Conv | 3 x 3/1 | 256 | 26 x 26 x256 |
| 22 | Conv | 1 x 1/1 | 255 | 26 x 26 x 255 |
| 23 | YOLO | | | |

A bounding box is characterized by the dimensions expressed as center coordinates $t_x$, $t_y$, width $t_w$ and height $t_h$. Moreover, each bounding box is accompanied by a confidence score $t_o$. This confidence score reflects how confident is the network that the box contains an object and also how accurate it thinks the box has the right dimension. More formally, as in the original paper of the first version of YOLO [19], we can define confidence as $Pr(Object) * IOU_{pred}^{truth}$ where $IOU_{pred}^{truth}$ is the intersection over unit between the ground truth box and the predicted one. Finally, each section has also C conditional class probabilities, $Pr(Class_i|Object)$ that in the YOLO layer are normalized with a sigmoid activation function. In the second version, the softmax activation function has been replaced because it assumes mutually exclusion between different classes. That is not true with larger data set like ImageNet [48], where labels are arranged as a directed graph [49] and not as flat structure.

Summarizing, the 1x1 convolution produces for each location an array of $[B * (5 + C)]$ elements where C is the number of classes of the training data set and B is the number of bounding boxes that for YOLOv3-tiny is equal to 3. So, for the COCO data set, C is equal to one, and so the output tensor of layer 15 in Tab. 2 is equal to 13 x 13 x 255.

It is essential to point out that, YOLO, starting from the second version, does not predict coordinates directly, but it refines hand-picked priors known as anchors. This approach simplifies the problem and makes it easier for the network to learn. Moreover, an ablation study made by the author [39] demonstrated how predicting offsets increased the recall of the network and stabilized gradients during training. So, for each cell, there are three bounding boxes with dimensions decided a priori. The network, during the learning phase, learns how to adjust these bounding boxes when an object is detected. Anchors dimension is critical, and in order not to manually guess their prior dimension, authors of the YOLO paper suggest to use k-mean clustering over the data set ground truth bounding boxes in order to predict the width and height of the box as offsets from cluster centroids. That allows generating prior bounding boxes that initialize the model with better representation and make the task easier to learn.

As presented in Fig. 4. the center of the object falls in a certain cell of the 13 x 13 grid. That cell is the designated one that has to detect the presence of the object. During training, we force the network to use the anchor box that has the prior dimension with the higher IOU with the ground truth. So, the network will generate $t_x$, $t_y$, $t_w$ and $t_h$ in order to refine one of its anchors present in that cell to perform the detection of the object. Then the YOLO layer will apply the following transformations in order to compute the dimension of the refined anchor box:

$$b_x = \sigma(t_x + c_x) \quad (1)$$
$$b_y = \sigma(t_y + c_y) \quad (2)$$
$$b_w = p_w e^{t_h} \quad (3)$$
$$b_h = p_h e^{t_h} \quad (4)$$

$b_x$, $b_y$, $b_w$ and $b_h$ are the x, y center coordinates, width and height of our prediction. On the other hand, $c_x$ and $c_y$ are instead the top-left coordinates of the grid designated for the prediction and $p_w$, $p_h$ are anchors prior dimensions of the box.

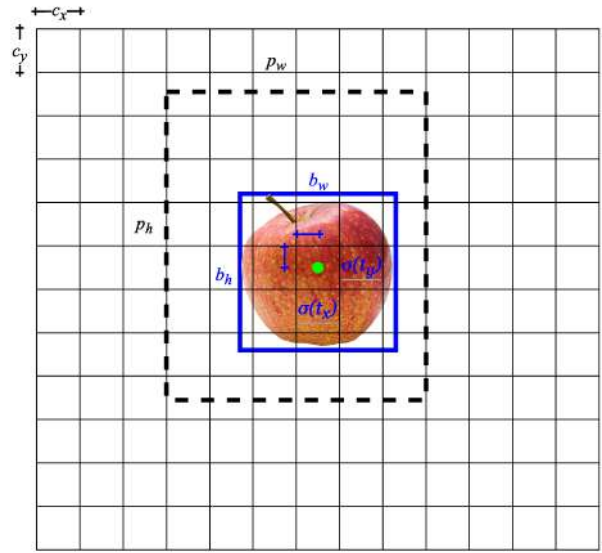

**FIGURE 4.** Yolov3-Tiny, as the region proposal network (RPN) [38] in Faster R-CNN, predicts the dimension of the bounding boxes as offsets to pre-defined hand-picked prior boxes known as anchors. Predicting the log-space transforms and not directly the dimension of the bounding boxes prevents unstable gradients during training and simplifies the problem making easier for the network to learn.

In conclusion, once trained, YOLOv3-tiny outputs a tensor of N x N x $[3 * (5 + C)]$. So, for each of N x N location the network outputs three bounding boxes with their related confidence $\sigma(t_o)$ and C class probabilities $p_c$. Then, confidence score and Non-maximum Suppression (NMS) are used to remove multiple detections and produce the final output bounding box predictions.

### C. A CUSTOM YOLOV3-TINY FOR SMALL OBJECTS DETECTION

The two first versions of YOLO greatly struggle at detecting small objects in the input image. That is mainly because that YOLO imposes a strong spatial constraint on bounding box predictions since each grid cell can only predict three bounding boxes. Moreover, the input dimension reduction performed in the backbone induces loss of low-level features, which are instrumental for detecting small objects. For this reason, the authors proposed with the third version of the "Tiny" network [20] an FPN structure producing detection at two different scales. The network downsamples the input raw image until the first detection layer, where a prediction is made using feature maps of a layer with stride 32. Then, layers are upsampled by a factor of two and concatenated with representation maps of a previous layer having identical feature map sizes.

We took this concept even further, and we build a YOLOv3-tiny architecture with predictions across three different scales. So, our model makes detection at feature maps of three different sizes, having strides 32, 16, and 8. Moreover, the input dimension has been increased from 416 × 416 to 608 × 608, and so, detections are made on scales 19 × 19, 38 × 38, and 76 × 76.

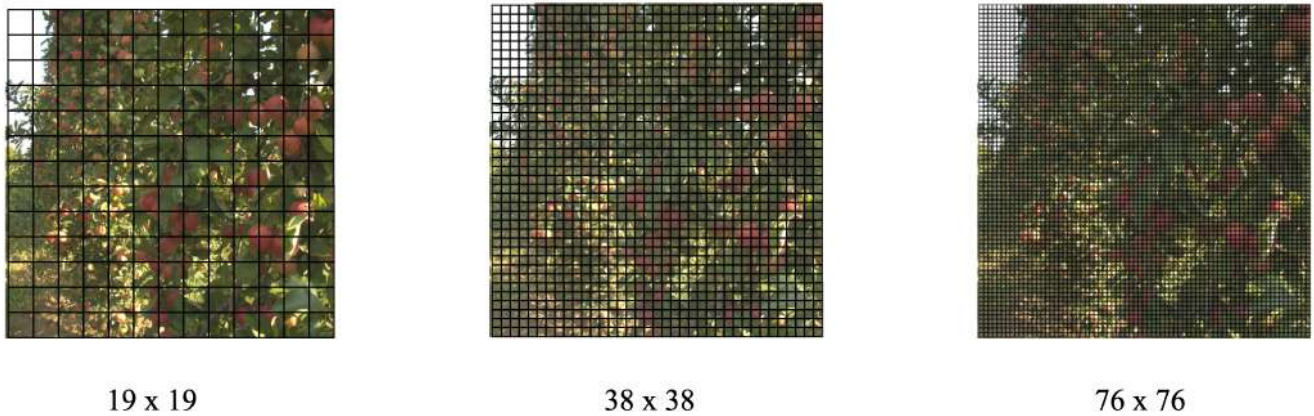

19 x 19    38 x 38    76 x 76

**FIGURE 5.** Our custom version of YOLOv3-tiny predicts three different scales. The detection layers make detection at features maps of three different sizes, having strides 32, 16, 8, respectively. So, having in our model an increased input size of 608 × 608, detections are made on scales 19 × 19, 38 × 38 and 76 × 76.

Fig. 3. presents the overall architecture of the modified network. It is clear from the representation, the pyramid structure: extracted features from the backbone are sampled at two different points and are lately concatenated with upsampled higher-level representations. In this way, high-level features are enriched with low-level information that helps the network learn fine-grained features that are fundamental for detecting small objects. At each scale, YOLO layers predict three bounding box per cell using three different anchors. So, the total number of anchors used is nine and the total number of bounding boxes predicted by the modified network is equal to [(19 x 19) * 3 + (38 x 38) * 3 + (76 × 76) * 3] = 22743. Confidence scores and later NMS will take care to drastically reduce this number removing low confidence boxes and redundant detections.

On Fig. 5. is depicted a graphical representation of the tensors used for the predictions at the three different scales. Each grid represents the tensor used for the prediction; overlapped with the original input, it shows where the model is making the prediction. It is noticeable how a finer grid can greatly improve the prediction capabilities of small objects. Indeed, for our specific case, this simple modification largely improves the accuracy of the model with a small overhead in terms of speed and power consumption.

## IV. EXPERIMENTAL RESULTS AND DISCUSSION

In this section, firstly some technical details of the training process of the modified architecture of YOLOv3-tiny are presented, then experimental evaluations are discussed for both the model and its deployment on the selected embedded platforms. Finally, quantitative and qualitative results are reported with a comparison between the different devices.

### A. EXPERIMENTAL TRAINING SETTINGS

We first processed raw data (stored in PNG format), introduced in section II-A, acquired from the test site in order to create a set of $n = 618$ samples images $\mathbb{X} = X_i$. Subsequently, inspired by Active learning techniques [50], in order to speed up label generation, we used an extensive and accurate version of YOLOv3 known as YOLOv3-spp, pre-trained on the COCO data set, to create a draft of the ground truth labels $\mathbb{Y} = y_i$. In order to increase the recall of the model, we set the network with a low value of confidence. Then, we performed manual inspection of the produced ground truth, using LabelMe [51], refining, and adjusting the predictions of the YOLOv3-spp network. Finally, we applied a straightforward pipeline to pre-process our set $\mathbb{X}$ of training samples; each image has been first resized from 5202×3465 at a 1920×1080 dimension and then, mean subtraction [52] has been applied to normalize the training data.

Using transfer learning, we started our training from a pre-trained backbone of the original model. That greatly speeds up the training, drastically reducing the number of training samples required to achieve a high level of accuracy. Except for the first two output dimensions, the backbone parameters are the same as the original model described in Table. 2. Indeed, having set the default input tensor to 608 × 608 × 3, the first two dimensions of the table are different for our modified implementation. Then, following the already discussed architecture in Fig. 3. the specifications for the number of filters, dimensions, and strides are presented in Table. 3. It is worth to notice the dimension of the prediction tensors at the three different scales; having only one class, the number of features is $[B*(5+C)] = 18$. This low number of dimensions further decreases the inference time required by the network.

We trained the network for 100 epochs using AdamW optimizer [53] and setting $\beta_1 = 0.89$ and $\beta_2 = 0.99$ and $\epsilon = 10^{-9}$.

$$\theta_{t+1} = \theta_t - \frac{\eta}{\sqrt{\hat{v}_t} + \epsilon} m_t - \eta w_d \theta_t \quad (5)$$

Moreover, AdamW updating rule (5) applies a slight modification to the original Adam updating rule; it subtracts a little portion of the weights at each step $\eta w_d \theta_t$ regularizing large weights inside the network and giving advantage to

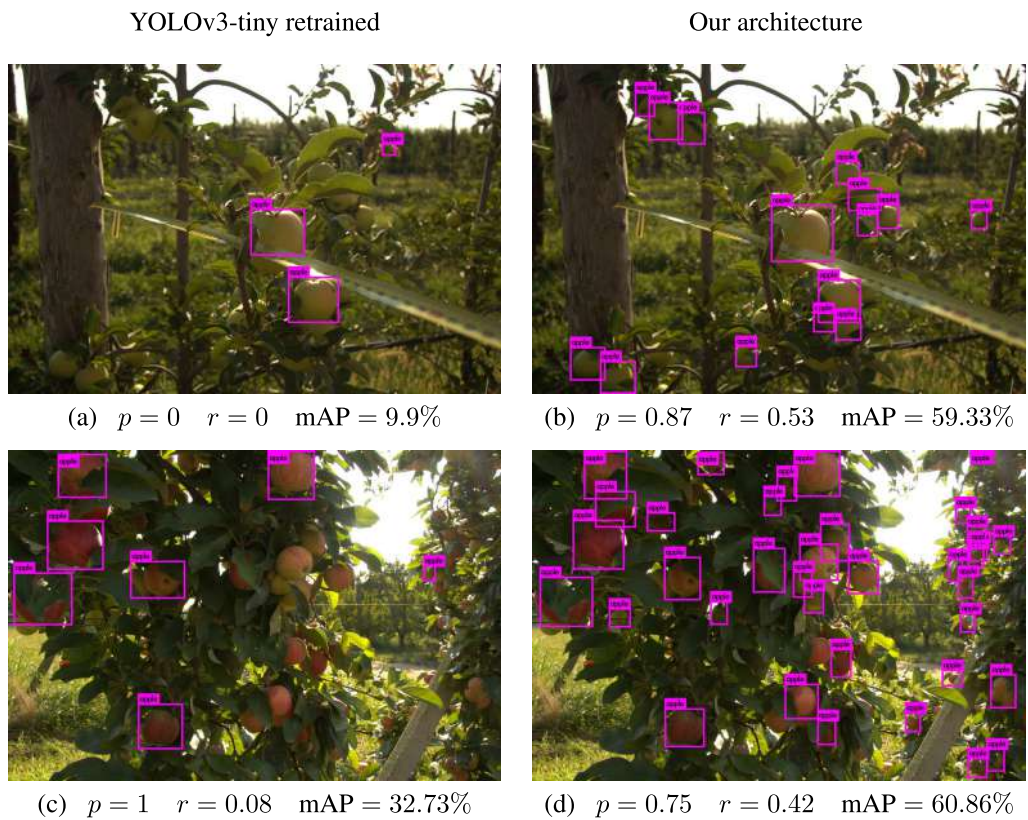

(a) $p = 0 \quad r = 0 \quad \text{mAP} = 9.9\%$

(b) $p = 0.87 \quad r = 0.53 \quad \text{mAP} = 59.33\%$

(c) $p = 1 \quad r = 0.08 \quad \text{mAP} = 32.73\%$

(d) $p = 0.75 \quad r = 0.42 \quad \text{mAP} = 60.86\%$

**FIGURE 6.** Comparison between the original YOLOv3-tiny and our modified version that performs predictions at three different scales. Fig. (b). and (d). produced by our custom version demonstrate a considerable improvement over the predictions of the original model. The additional detection at a scale with stride 8 largely improve the recall of the model and make it much more robust to scale variations. All the predcition are done with default confidence threshold $c = 0.25$. Precision and recall are computed for $IOU_{\text{target}} = 0.5$.

**TABLE 3.** FPN section of our YOLOv3-tiny modified architecture. Detections are made with feature maps at three different scales, obtained fusing upsampled high level features with low level representations of the same size.

| Layer | Type | Size/Stride | Filters | Output |
|---|---|---|---|---|
| 13 | Conv | 1 x 1/1 | 256 | 19 x 19 x 256 |
| 14 | Conv | 3 x 3/1 | 512 | 19 x 19 x 512 |
| 15 | Conv | 1 x 1/1 | 255 | 19 x 19 x 18 |
| 16 | YOLO | | | |
| 17 | Route 13 | | | |
| 18 | Conv | 1 x 1/1 | 128 | 19 x 19 x 128 |
| 19 | Up-samp | 2 x 2/1 | | 38 x 38 x 128 |
| 20 | Route 19 8 | | | |
| 21 | Conv | 3 x 3/1 | 256 | 38 x 38 x256 |
| 22 | Conv | 1 x 1/1 | 18 | 38 x 38 x 18 |
| 23 | YOLO | | | |
| 24 | Route 21 | | | |
| 25 | Conv | 1 x 1/1 | 128 | 38 x 38 x 128 |
| 26 | Up-samp | 2 x 2/1 | | 76 x 76 x 128 |
| 27 | Route 26 6 | | | |
| 28 | Conv | 3 x 3/1 | 256 | 76 x 76 x 256 |
| 29 | Conv | 1 x 1/1 | 18 | 76 x 76 x 18 |
| 30 | YOLO | | | |

rare features. In all our experimental evaluations it proved to give better results than classical L2 regularization [54] that modifies directly the cost function. We set $w_d = 0.0001$.

All training has been carried out on a workstation with an NVIDIA RTX 2080 Ti and 64GB of DDR4 SDRAM. The training took on average one-hour using the TensorFlow framework and CUDA 10.

### B. QUANTITATIVE RESULTS: MODEL PERFORMANCE

To understand the model performance, mean average precision (mAP) is computed on the test dataset. Mean average precision is a popular object detection scoring method that assesses the network performance in detecting the target objects for different values of target intersection over unit $IOU_{\text{target}}$. This methodology has been presented for the PASCAL Visual Object Classification (VOC) challenge 2012 [55]. Each predicted bounding box $i$ is compared with the ground truth and marked as correct (true positive TP) if the apple is present and $IOU_i > IOU_{\text{target}}$. If $IOU_i$ is lower than the target or the apple is not present, the prediction is marked as incorrect (false positive FP). Finally, all the apples not detected are marked as missing predictions (false negative FN). Since the predicted bounding boxes are given as output only if they have a level of confidence above a certain threshold $c$, it is possible to compute the precision ($p$) and the recall ($r$) of the network over the test dataset as a function of $c$:

$$p(c) = \frac{\text{TP}(c)}{\text{TP}(c) + \text{FP}(c)} \qquad r(c) = \frac{\text{TP}(c)}{\text{TP}(c) + \text{FN}(c)} \quad (6)$$

**TABLE 4.** Detection performance of the network on the test dataset from OIDv4. The same computation is made with the original YOLOv3-tiny architecture retrained on the same training dataset. The results show how the proposed architecture can boost the recognition in agriculture context by allowing small fruits detection.

| $IOU_{target}$ | Proposed architecture | | | Original Yolov3-Tiny | Gain |
|---|---|---|---|---|---|
| | Recall ($c = 0.25$) | Precision ($c = 0.25$) | mAP | mAP | |
| 0.50 | 0.83 | 0.69 | 83.64% | 77.02 % | 6.62 % |
| 0.75 | 0.55 | 0.46 | 47.97% | 42.50 % | 5.47 % |

**TABLE 5.** Comparison between different devices power consumption and performances achieved with our customized version of YOLOv3-tiny. Jetson series boards can be run at different power modes reducing current absorption at the expense of lowering computational capabilities. The mode column shows the theoretical maximum absorbed power in the different working modality, that is different from the actual dissipated power during the execution of the algorithm. The best performance, in terms of frame per second, is highlighted with a red rectangle.

| Device | Mode | $V_{al}[V]$ | $I_{mean}[A]$ | $P[W]$ | fps |
|---|---|---|---|---|---|
| Jetson AGX Xavier | IDLE 30W | 19 | 0.35 | 6.65 | N.A. |
| | RUNNING 30W | 19 | 1 | 19 | 30 |
| | RUNNING 15W | 19 | 0.88 | 16.72 | 25 |
| | RUNNING 10W | 19 | 0.66 | 12.54 | 13 |
| Jetson Nano | IDLE 10W | 5 | 0.32 | 1.6 | N.A. |
| | RUNNING 10W | 5 | 2.04 | 10.20 | 8 |
| | RUNNING 5W | 5 | 1.42 | 7.1 | 6 |
| Raspberry Pi 3B+ | IDLE | 5 | 0.61 | 3.075 | N.A. |
| RP3 + Neural Stick 1 | RUNNING | 5 | 1.2 | 6 | 4 |
| RP3 + Neural Stick 2 | RUNNING | 5 | 1.12 | 5.6 | 5 |

Computing all the possible values of $p(c)$ and $r(c)$, it is possible to get the precision/recall curve. The graph is then usually smoothed in order to get a monotonically decreasing precision curve by setting $p(r) = \max_{r' \geq r} p(r')$. The average precision of the network is computed as the area under the obtained curve and is always a number between 0 and 1:

$$AP = \int_0^1 p(r)dr \qquad (7)$$

An average precision equal to 1 means that the detector is able to reach a perfect precision (100%) for all the values of recall. Thus it is possible to find a value of $c$ such that we are able to detect all the objects with correct bounding boxes. On the other hand, an average precision of 0 means that we cannot detect any object correctly whatever value of $c$ we choose, thus both $p(c)$ and $r(c)$ are always equal to 0.

For a multi-class object detection algorithm, the mean average precision is the mean of the AP over all the classes. In our specific context, we are dealing with apples only, thus AP = mAP. The mean average precision gives thus a piece of information on the quality of the network detection independent from the chosen $c$, that can be chosen considering what is more important among precision and recall for the specific application.

Different values of mAP can be computed depending on the selected $IOU_{target}$. Usual values are 0.5 and 0.75 in order to evaluate the model performance with different requirements on the detection accuracy of the locations of the objects. Table. 4. presents the recall and precision for the default confidence threshold $c = 0.25$ and the mean average precision for the two values of $IOU_{target}$. The same computation has been performed with the original YOLOv3-tiny architecture, retrained for the apple detection only with the same methodology described in section IV-A. The results presented in Table. 4. show how the change in the architecture can boost the mAP on the test dataset of up to 6.6%.

### C. QUANTITATIVE RESULTS: EMBEDDED IMPLEMENTATION

After the training, the model has been deployed on the different hardware platforms presented in section II-B. We tested the performance in terms of absorbed power and frame rate.

Firstly we measured the power consumption of the different boards (Jetson AGX Xavier, Jetson Nano, Raspberry Pi 3B+) at idle condition, and then we executed the algorithm for nearly 5 minutes to be sure to be at steady state. We measured directly the current absorbed from the power source, thus obtaining the power consumption of the entire system. Since the Jetson boards allow the user to select different working conditions, we tested all of them. The results are presented in Table. 5.

The NVIDIA Jetson AGX Xavier is the most performing platform, being able to reach 30 fps in the 30W operational mode. Also, in the other modalities, it is able to reach frame rates suitable for strong real team applications. With the Jetson Nano, the frame rate drops to 8 fps in 10W mode, which can still be an acceptable value for soft real-time contexts. With the Raspberry Pi and the Intel's NCS, the performance is further lowered. With the same running conditions, the more advanced NCS2 is able to outperform its predecessor both in terms of frame rate and power consumption. However, despite being more flexible, these USB accelerators cannot go beyond the five fps in the best case.

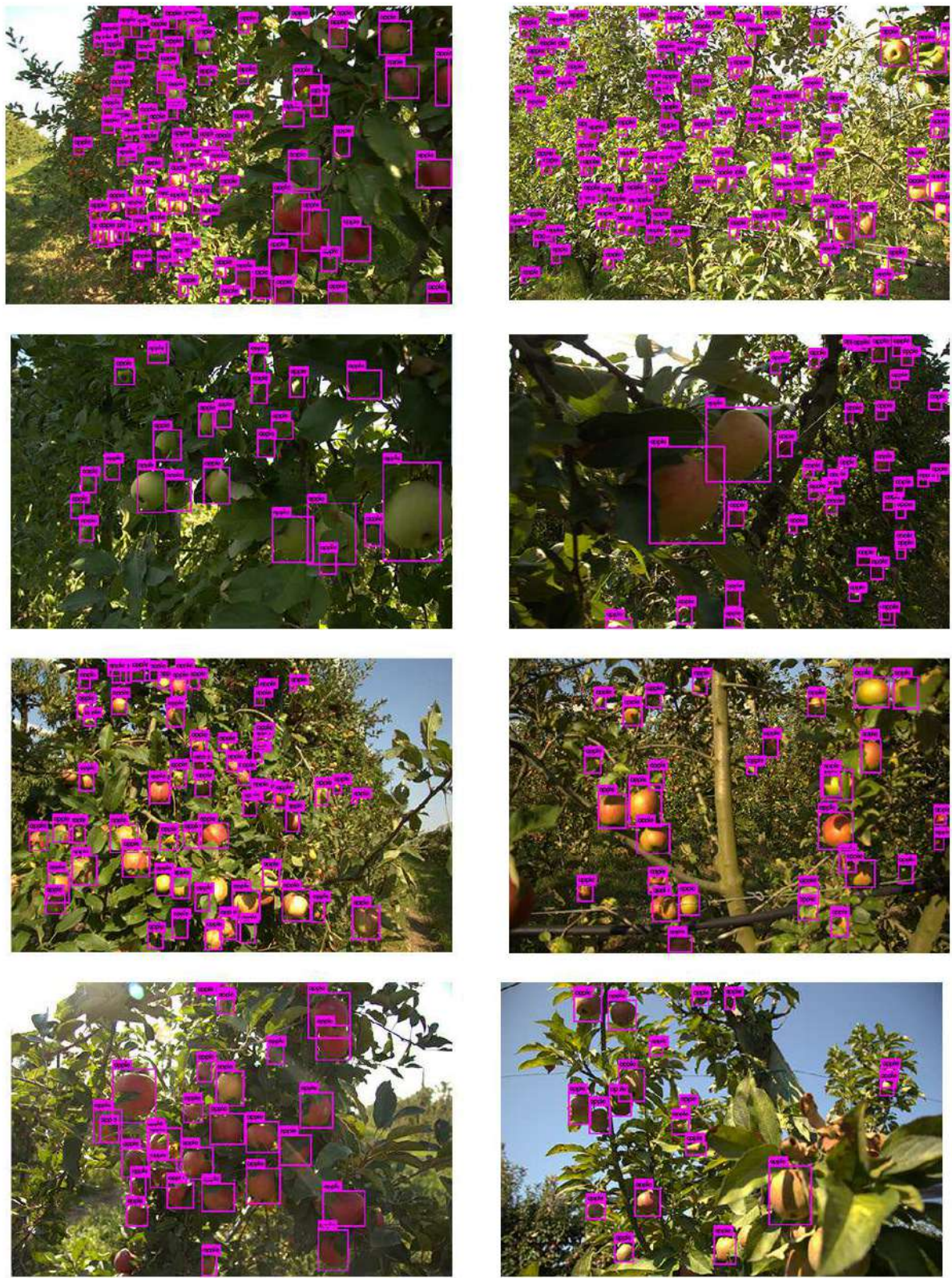

**FIGURE 7.** Qualitative results of some additional test images acquired from the same study site of the training data set. It is possible to notice how our custom version of YOLOv3-tiny is robust to different factors of variation [56] in how apples appear. Simultaneously handling variations in illumination, viewpoint, scale, occlusion, and background clutter is a challenging task that our system has to tackle in real-time with limited computational capabilities.

An interesting comparison between the different platforms is the price/fps ratio, shown in Table. 6. The Jetson Nano appears to be the best choice if we are looking for a balance between performance and cost. On the other hand, the AGX

TABLE 6. Price/performance analysis of the different embedded platforms. The best price/fps ratio is highlighted with a red rectangle.

| Device | price | $fps_{max}$ | price/fps |
|---|---|---|---|
| Intel Movidius NCS | 70 $ | 4 | 17.5 |
| Intel NCS 2 | 74 $ | 5 | 14.8 |
| Jetson Nano | 99 $ | 8 | 12.38 |
| Jetson AGX Xavier | 800 $ | 30 | 26.67 |

Xavier has the higher ratio, since it is a board with the highest quality, but certainly not suitable for low-cost solutions. The Intel Neural Sticks results in the second and third place for price/fps ratio, but it must be underlined that, since they are USB accelerators only, an additional embedded computer must be purchased, increasing the final cost.

### D. QUALITATIVE RESULTS AND COMPARISON

Fig. 7 presents some test images excluded during the training phase. It is possible to see how our network is able to recognize a great number of apples in several image conditions such as different illumination and contrast. The network is able to detect the fruits on different scales, and in particular it can recognize very small apples, even in bad lighting conditions.

A comparison with the original architecture is presented in Fig. 6. We processed two images excluded from the training dataset with default confidence threshold $c = 0.25$ and we computed precision and recall with $IOU_{target} = 0.5$. It's interesting to notice that precision and recall for image (a) are both 0, since no bounding box have sufficient intersection over unit to be considered as a true positive. Image (c) has precision equal to 1, but very poor recall, since the network is able to detect only the 8% of the apples. Our architecture, on the other hand, is able to strongly increase the recall detecting very small fruits, boosting the quality of the predictions. In this scenario, the mAP gain is a lot higher with respect to the test dataset taken from OIDv4. This is due to the fact that the images of the apple class in the OIDv4 dataset present, on average, bigger apples with respect to the training dataset taken on a real orchard, so the difference between the two architectures is less visible. On the other hand, on the test images taken from the same dataset used for training, the ability of detecting little apples becomes fundamental to reach a high recall value. However, we presented our results for the OIDv4 in order to make the experiments repeatable and allow direct comparison with our work.

### V. CONCLUSION

A real-time apple detection system has been developed and tested on several edge AI devices. The classical YOLOv3-tiny architecture has been modified and adapted in order to increase its accuracy in the presence of small and largely occluded objects. It has been trained with a custom data set, acquired on a real orchard, and tested with all available images of OIDv4. Accuracy results have demonstrated a boost in terms of recall and precision in the presence of targets with disparate sizes. Experimental evaluations have been carried out in order to highlight performances achieved in terms of inference speed and power consumption by the different embedded solutions selected. Experimentation results have shown promising prospects to exploit the tested system to produce real-time positions and numbers of detections with minimal power consumption. A complete framework could integrate the presented research for diverse purposes, from apple counting, harvesting health assessment to smart packaging. Indeed, further works will only focus on yield estimation using the proposed methodology to count the number of apples reliably. Indeed, image registration has not been directly addressed in this study, but it is a strong requirement in order to reduce double count and improve the precision of the system. Moreover, in view of an extension of the presented analysis, FPGA/ASIC implementation will be considered for future developments of this research study. Finally, the adopted methodology is not limited to apple detection task, but could also be implemented for other applications where the detection of small and tiny objects in real-time at the edge is needed.

### REFERENCES

[1] O. Cohen, R. Linker, and A. Naor, "Estimation of the number of apples in color images recorded in orchards," in *Proc. Int. Conf. Comput. Comput. Technol. Agricult.*, 2011, pp. 630–642.

[2] W. Ji, D. Zhao, F. Cheng, B. Xu, Y. Zhang, and J. Wang, "Automatic recognition vision system guided for apple harvesting robot," *Comput. Elect. Eng.*, vol. 38, no. 5, pp. 1186–1195, Sep. 2012.

[3] Z. De-An, L. Jidong, J. Wei, Z. Ying, and C. Yu, "Design and control of an apple harvesting robot," *Biosyst. Eng.*, vol. 110, no. 2, pp. 112–122, Oct. 2011.

[4] A. Gongal, A. Silwal, M. Karkee, Q. Zhang, K. Lewis, and S. Amatya, "Apple crop-load estimation with over-the-row machine vision system," *Comput. Electron. Agricult.*, vol. 120, pp. 26–35, Jan. 2016.

[5] A. Silwal, A. Gongal, and M. Karkee, "Apple identification in field environment with over the row machine vision system," *Agricult. Eng. Int., CIGR J.*, vol. 16, no. 4, pp. 66–75, 2016.

[6] Q. Wang, S. Nuske, M. Bergerman, and S. Singh, "Automated crop yield estimation for apple orchards," in *Experimental Robotics*. Heidelberg, Germany: Springer, 2013, pp. 745–758.

[7] Z. Wang, K. Walsh, and B. Verma, "On-tree mango fruit size estimation using RGB-D images," *Sensors*, vol. 17, no. 12, p. 2738, Nov. 2017.

[8] D. Stajnko, M. Lakota, and M. Hočevar, "Estimation of number and diameter of apple fruits in an orchard during the growing season by thermal imaging," *Comput. Electron. Agricult.*, vol. 42, no. 1, pp. 31–42, Jan. 2004.

[9] M. Regunathan and W. Suk Lee, "Citrus fruit identification and size determination using machine vision and ultrasonic sensors," in *Proc. ASAE Annu. Meeting*, Tampa, FL, USA, Jul. 2005.

[10] A. Kamilaris and F. X. Prenafeta-Boldú, "Deep learning in agriculture: A survey," *Comput. Electron. Agricult.*, vol. 147, pp. 70–90, Apr. 2018.

[11] J. L. Tang, D. Wang, Z. G. Zhang, L. J. He, J. Xin, and Y. Xu, "Weed identification based on K-means feature learning combined with convolutional neural network," *Comput. Electron. Agricult.*, vol. 135, pp. 63–70, Apr. 2017.

[12] Y.-D. Zhang, Z. Dong, X. Chen, W. Jia, S. Du, K. Muhammad, and S. H. Wang, "Image based fruit category classification by 13-layer deep convolutional neural network and data augmentation," *Multimedia Tools Appl.*, to be published.

[13] P. A. Dias, A. Tabb, and H. Medeiros, "Apple flower detection using deep convolutional networks," *Comput. Ind.*, vol. 99, pp. 17–28, Aug. 2018.

[14] S. Bargoti and J. Underwood, "Deep fruit detection in orchards," in *Proc. IEEE Int. Conf. Robot. Automat. (ICRA)*, 2017, pp. 3626–3633.

[15] K. Yamamoto, W. Guo, Y. Yoshioka, and S. Ninomiya, "On plant detection of intact tomato fruits using image analysis and machine learning methods," *Sensors*, vol. 14, no. 7, pp. 12191–12206, 2014.

[16] S. W. Chen, S. S. Skandan, S. Dcunha, J. Das, E. Okon, C. Qu, C. J. Taylor, and V. Kumar, "Counting apples and oranges with deep learning: A data-driven approach," *IEEE Robot. Autom. Lett.*, vol. 2, no. 2, pp. 781–788, Jan. 2017.

[17] S. Ren, K. He, R. Girshick, and J. Sun, "Faster R-CNN: Towards real-time object detection with region proposal networks," *IEEE Trans. Pattern Anal. Mach. Intell.*, vol. 39, no. 6, pp. 1137–1149, Jun. 2017.

[18] K. Simonyan and A. Zisserman, "Very deep convolutional networks for large-scale image recognition," 2014, *arXiv:1409.1556*. [Online]. Available: https://arxiv.org/abs/1409.1556

[19] J. Redmon, S. Divvala, R. Girshick, and A. Farhadi, "You only look once: Unified, real-time object detection," in *Proc. IEEE Conf. Comput. Vis. Pattern Recognit.*, Jun. 2016, pp. 779–788.

[20] J. Redmon and A. Farhadi, "YOLOv3: An incremental improvement," 2018, *arXiv:1804.0276*. [Online]. Available: https://arxiv.org/abs/1804.0276

[21] Y. You, Z. Zhang, C.-J. Hsieh, J. Demmel, and K. Keutzer, "ImageNet training in minutes," in *Proc. 47th Int. Conf. Parallel Process. (ICPP)*, 2018.

[22] M. Duan, K. Li, X. Liao, and K. Li, "A parallel multiclassification algorithm for big data using an extreme learning machine," *IEEE Trans. Neural Netw. Learn. Syst.*, vol. 29, no. 6, pp. 2337–2351, Jun. 2018.

[23] A. Zhou, A. Yao, Y. Guo, L. Xu, and Y. Chen, "Incremental network quantization: Towards lossless cnns with low-precision weights," 2017, *arXiv:1702.03044*. [Online]. Available: https://arxiv.org/abs/1702.03044

[24] M. Rastegari, V. Ordonez, J. Redmon, and A. Farhadi, "XNOR-Net: ImageNet classification using binary convolutional neural networks," in *Proc. Eur. Conf. Comput. Vis.*, 2016, pp. 525–542.

[25] X. Xu, J. Amaro, S. Caulfield, G. Falcao, and D. Moloney, "Classify 3D voxel based point-cloud using convolutional neural network on a neural compute stick," in *Proc. 13th Int. Conf. Natural Comput., Fuzzy Syst. Knowl. Discovery (ICNC-FSKD)*, Jul. 2017, pp. 37–43.

[26] S. Mittal and J. S. Vetter, "A survey of CPU-GPU heterogeneous computing techniques," *ACM Comput. Surv.*, vol. 47, no. 4, pp. 1–35, Jul. 2015.

[27] C.-K. Lai, C.-W. Yeh, C.-H. Tu, and S.-H. Hung, "Fast profiling framework and race detection for heterogeneous system," *J. Syst. Archit.*, vol. 81, pp. 83–91, Nov. 2017.

[28] B. Reddy, Y.-H. Kim, S. Yun, C. Seo, and J. Jang, "Real-time driver drowsiness detection for embedded system using model compression of deep neural networks," in *Proc. IEEE Conf. Comput. Vis. Pattern Recognit. Workshops (CVPRW)*, Jul. 2017, pp. 121–128.

[29] S. Gu, X. Chen, W. Zeng, and X. Wang, "A deep learning tennis ball collection robot and the implementation on NVIDIA Jetson TX1 board," in *Proc. IEEE/ASME Int. Conf. Adv. Intell. Mechatronics (AIM)*, Jul. 2018, pp. 170–175.

[30] D. Kang, D. Kang, J. Kang, S. Yoo, and S. Ha, "Joint optimization of speed, accuracy, and energy for embedded image recognition systems," in *Proc. Design, Autom. Test Eur. Conf. Exhib. (DATE)*, Mar. 2018, pp. 715–720.

[31] S. Luo, H. Lu, J. Xiao, Q. Yu, and Z. Zheng, "Robot detection and localization based on deep learning," in *Proc. Chin. Autom. Congr. (CAC)*, Oct. 2017, pp. 7091–7095.

[32] I. Sa, Z. Chen, M. Popovic, R. Khanna, F. Liebisch, J. Nieto, and R. Siegwart, "WeedNet: Dense semantic weed classification using multispectral images and MAV for smart farming," *IEEE Robot. Autom. Lett.*, vol. 3, no. 1, pp. 588–595, Jan. 2018.

[33] NVIDIA. (2019). *Graphics Reinvented: NVIDIA GeForce RTX 2080 Ti Graphics Card.* [Online]. Available: https://www.nvidia.com/en-us/geforce/graphics-cards/rtx-2080-ti/

[34] Intel Software. (2019). *Intel Movidius Neural Compute Stick.* [Online]. Available: https://software.intel.com/en-us/movidius-ncs

[35] Intel Software. (2019). *Intel Neural Compute Stick 2.* [Online]. Available: https://software.intel.com/en-us/neural-compute-stick

[36] NVIDIA Developer. (2019). *NVIDIA Jetson—Hardware For Every Situation.* [Online]. Available: https://developer.nvidia.com/embedded/develop/hardware

[37] T.-Y. Lin, P. Goyal, R. Girshick, K. He, and P. Dollar, "Focal loss for dense object detection," in *Proc. IEEE Int. Conf. Comput. Vis.*, Oct. 2017, pp. 2980–2988.

[38] R. Girshick, "Fast R-CNN," in *Proc. IEEE Int. Conf. Comput. Vis. (ICCV)*, Dec. 2015, pp. 1440–1448.

[39] J. Redmon and A. Farhadi, "YOLO9000: Better, faster, stronger," in *Proc. IEEE Conf. Comput. Vis. Pattern Recognit. (CVPR)*, Jul. 2017, pp. 7263–7271.

[40] K. He, X. Zhang, S. Ren, and J. Sun, "Deep residual learning for image recognition," in *Proc. IEEE Conf. Comput. Vis. Pattern Recognit. (CVPR)*, Jun. 2016, pp. 770–778.

[41] G. Huang, Z. Liu, L. V. D. Maaten, and K. Q. Weinberger, "Densely connected convolutional networks," in *Proc. IEEE Conf. Comput. Vis. Pattern Recognit. (CVPR)*, Jul. 2017, pp. 4700–4708.

[42] B. Xu, N. Wang, T. Chen, and M. Li, "Empirical evaluation of rectified activations in convolutional network," 2015, *arXiv:1505.00853*. [Online]. Available: https://arxiv.org/abs/1505.00853

[43] S. Ioffe and C. Szegedy, "Batch normalization: Accelerating deep network training by reducing internal covariate shift," 2015, *arXiv:1502.03167*. [Online]. Available: https://arxiv.org/abs/1502.03167

[44] T.-Y. Lin, P. Dollar, R. Girshick, K. He, B. Hariharan, and S. Belongie, "Feature pyramid networks for object detection," in *Proc. IEEE Conf. Comput. Vis. Pattern Recognit. (CVPR)*, Jul. 2017, pp. 2117–2125.

[45] T.-Y. Lin, M. Maire, S. Belongie, J. Hays, P. Perona, D. Ramanan, P. Dollár, and C. L. Zitnick, "Microsoft COCO: Common objects in context," in *Proc. Eur. Conf. Comput. Vis.* Cham, Switzerland: Springer, 2014, pp. 740–755.

[46] A. Kuznetsova, H. Rom, N. Alldrin, J. Uijlings, I. Krasin, J. Pont-Tuset, S. Kamali, S. Popov, M. Malloci, T. Duerig, and V. Duerig, "The open images dataset v4: Unified image classification, object detection, and visual relationship detection at scale," 2018, *arXiv:1811.00982*. [Online]. Available: https://arxiv.org/abs/1811.00982

[47] M. Long, Y. Cao, J. Wang, and M. I. Jordan, "Deep transfer learning with joint adaptation networks," in *Proc. Int. Conf. Mach. Learn.*, vol. 70, 2017, pp. 2208–2217.

[48] J. Deng, W. Dong, R. Socher, L.-J. Li, K. Li, and L. Fei-Fei, "ImageNet: A large-scale hierarchical image database," in *Proc. IEEE Conf. Comput. Vis. Pattern Recognit.*, Jun. 2009, pp. 248–255.

[49] C. Fellbaum, Ed., *WordNet: An Electronic Lexical Database*. Cambridge, MA, USA: MIT Press, 1998.

[50] Y. Fu, X. Zhu, and B. Li, "A survey on instance selection for active learning," *Knowl. Inf. Syst.*, vol. 35, no. 2, pp. 249–283, May 2013.

[51] B. C. Russell, A. Torralba, K. P. Murphy, and W. T. Freeman, "LabelMe: A database and Web-based tool for image annotation," *Int. J. Comput. Vis.*, vol. 77, nos. 1–3, pp. 157–173, May 2008.

[52] T. DeVries and G. W. Taylor, "Improved regularization of convolutional neural networks with cutout," 2017, *arXiv:1708.04552*. [Online]. Available: https://arxiv.org/abs/1708.04552

[53] I. Loshchilov and F. Hutter, "Fixing weight decay regularization in adam," 2017, *arXiv:1711.05101*. [Online]. Available: https://arxiv.org/abs/1711.05101

[54] R. Moore and J. DeNero, "L1 and L2 regularization for multiclass hinge loss models," in *Proc. Symp. Mach. Learn. Speech Lang. Process.*, 2011.

[55] M. Everingham, L. Van Gool, C. K. I. Williams, J. Winn, and A. Zisserman, "The Pascal visual object classes (VOC) challenge," *Int. J. Comput. Vis.*, vol. 88, no. 2, pp. 303–338, Jun. 2010.

[56] A. Rosebrock, *Deep Learning for Computer Vision With Python: Starter Bundle*. Heidelberg, Germany: PyImageSearch, 2017.

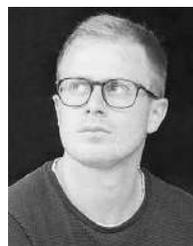

**VITTORIO MAZZIA** received the master's degree in mechatronics engineering from the Politecnico di Torino, presenting a thesis "Use of deep learning for automatic low-cost detection of cracks in tunnels," developed in collaboration with California State University. He is currently pursuing the Ph.D. degree in electrical, electronics, and communications engineering with the two interdepartmental centres PIC4SeR and SmartData. His current research interests involve deep learning applied to different tasks of computer vision, autonomous navigation for service robotics, and reinforcement learning. Moreover, making use of neural compute devices (like Jetson Xavier, Jetson Nano, Movidius Neural Stick) for hardware acceleration, he is currently working on machine learning algorithms and their embedded implementation for AI at the edge.

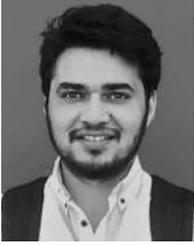

**ALEEM KHALIQ** received the B.S. and M.S. degrees in electronic engineering from International Islamic University Islamabad (IIUI), Pakistan, in 2008 and 2012, respectively. He is currently pursuing the Ph.D. degree from the Electrical, Electronics, and Communications Engineering Program, Politecnico di Torino, Italy. Since 2010, he has been a Lab Engineer (currently on study leave) with the Department of Electrical Engineering, Faculty of Engineering and Technology, IIUI. He is also an Active Member of Polito Inter-departmental Center for Service Robotics (PIC4SeR). His research interests include remote sensing applications in agriculture, applications of machine learning in satellite imagery, optical satellite and aerial imagery, and image processing. He received the scholarship from the Higher Education Commission of Pakistan to pursue his Ph.D. studies in abroad.

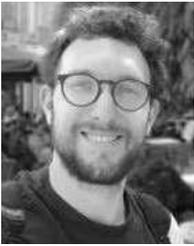

**FRANCESCO SALVETTI** received the bachelor's degree in electronic engineering and the master's degree in mechatronics engineering from the Politecnico di Torino, in 2017 and 2019, respectively. He is currently pursuing the Ph.D. degree in electrical, electronics, and communications engineering in collaboration with the two interdepartmental centres PIC4SeR and Smart Data at the Politecnico di Torino, Italy. He is currently working on machine learning applied to computer vision and image processing in robotics applications.

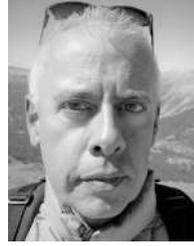

**MARCELLO CHIABERGE** is currently an Associate Professor with the Department of Electronics and Telecommunications, Politecnico di Torino, Turin, Italy. He is also the Co-Director of the Mechatronics Lab, Politecnico di Torino, Turin, and the Director and the Principal Investigator of the new Centre for Service Robotics (PIC4SeR), Turin. He has authored more than 100 articles accepted in international conferences and journals and the coauthor of nine international patents. His research interests include hardware implementation of neural networks and fuzzy systems and the design and implementation of reconfigurable real-time computing architectures.

● ● ●